\documentclass[lettersize,journal]{IEEEtran}
\usepackage{amsmath,amsfonts}
\usepackage{algorithmic}
\usepackage{algorithm}
\usepackage{array}
\usepackage[caption=false,font=normalsize,labelfont=sf,textfont=sf]{subfig}
\usepackage{textcomp}
\usepackage{stfloats}
\usepackage{url}
\usepackage{verbatim}
\usepackage{graphicx}
\usepackage{cite}
\usepackage{color}
\usepackage{multirow}
\def\degree{${}^{\circ}$}
\usepackage[table,xcdraw]{xcolor}
\usepackage{array}
\usepackage{arydshln}
\usepackage{enumerate}
\usepackage{caption}
\usepackage[normalem]{ulem}
\useunder{\uline}{\ul}{}
\usepackage[table,xcdraw]{xcolor}
\usepackage{booktabs}

\newcommand{\etal}{\textit{et al}. }

\usepackage{array}
\makeatletter
\newcommand{\thickhline}{%
    \noalign {\ifnum 0=`}\fi \hrule height 0.7pt
    \futurelet \reserved@a \@xhline
}
\newcolumntype{"}{@{\hskip\tabcolsep\vrule width 0.6pt\hskip\tabcolsep}}

\begin{document}

\title{Attention in Attention Network for Image Super-Resolution}

\author{Haoyu Chen, Jinjin Gu, Zhi Zhang

}

\maketitle

\begin{abstract}
Convolutional neural networks have allowed remarkable advances in single image super-resolution (SISR) over the last decade.
Among recent advances in SISR, attention mechanisms are crucial for high performance SR models.
However, attention mechanism remains unclear on why and how it works in SISR task.
In this work, we attempt to quantify and visualize attention mechanisms in SISR and show that not all attention modules are equally beneficial.
We then propose attention in attention network (A$^2$N) for more efficient and accurate SISR.
Specifically, A$^2$N consists of a non-attention branch and a coupling attention branch. 
A dynamic attention module is proposed to generate weights for these two branches to suppress unwanted attention adjustments dynamically, where the weights could change adaptively according to the input features. 
This allows attention modules to specialize to beneficial examples without otherwise penalties and thus greatly improve the capacity of the attention network with few parameters overhead.
Experimental results demonstrate that our final model — A\textsuperscript{2}N could achieve superior trade-off performances comparing with state-of-the-art networks of similar sizes.
\end{abstract}


\section{Introduction}

\IEEEPARstart{I}{mage} super-resolution (SR) is a low-level computer vision problem, which aims at recovering a high-resolution (HR) image from a low-resolution (LR) observation. 
In recent years, SR methods based on deep convolution neural networks (CNNs) have achieved significant success.
Recently, advanced methods begin to aggregate attention mechanism into the SR model, \textit{e.g.}, channel attention and spatial attention \cite{zhang2018image,hu2019channel,kim2018ram,wang2020lightweight,ma2020accurate}.
The introduction of attention mechanisms greatly improves the performance of these networks and some works \cite{zhang2018image,behjati2020hierarchical} argue that this improvement is due to the re-calibration of the feature responses towards the most informative and important components of the inputs.
We naturally raise two questions: 1) what kind of features would attention mechanisms response to? 2) is it always beneficial to enhance these features?

In this paper, we answer the first question by assessing the most valuable areas highlighted by attention.
We observed that for an SR network, the attention modules in the early layers of the network tend to enhance the low-frequency bands of the features, 
the blocks in-between have mixed responses,
and the end attention modules enhance the high-frequency components of the feature maps such as edges and textures.
For the second question, the ablation experiments on the attention module reveal interesting conclusions.
Only conducting attention operations at early layers brings very limited improvement, and ablating these modules that focus on low frequencies can even get better performance.
This shows that not all attention modules have the same contribution, the use of some attention modules can even cause performance drops.
Based on the above findings, we propose a novel low-consumption dynamic attention module called attention in attention (A$^2$) structure, which is divided into attention branch and non-attention branch.
The attention branch is used to enhance useful information, and the non-attention branch aims to learn beneficial information that is ignored by attention modules.
We propose a dynamic attention module to dynamically allocate the weights of two branches, make full use of the information of the two branches, enhance the high contribution information and suppress the redundant information.
This module works in both training and inference phases.
Experiments have proved that the A$^2$ structure is better than the traditional attention structure.

\begin{figure}[t]
\begin{center}
   \includegraphics[width=0.99\linewidth]{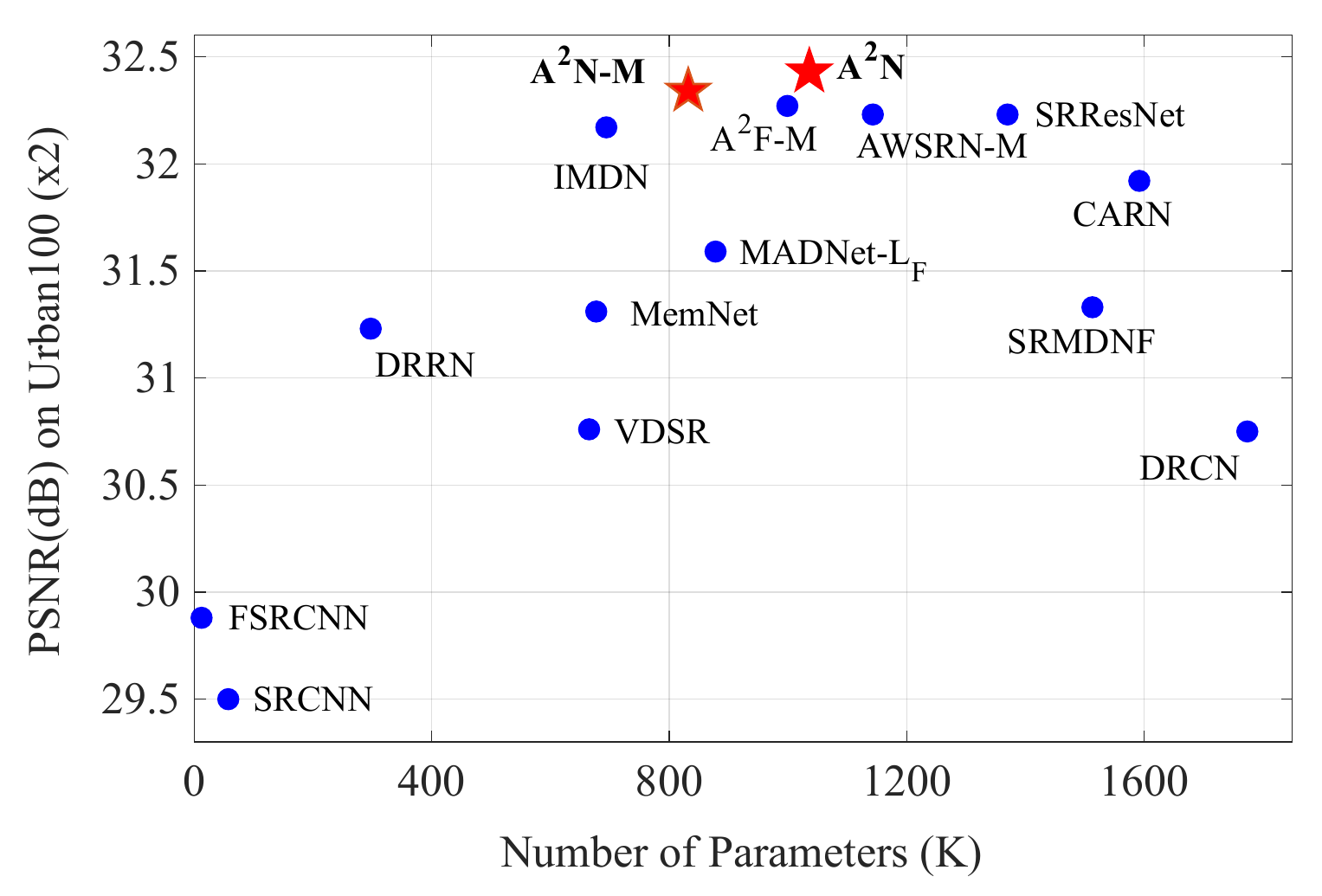}
\end{center}
\caption{Performance comparison between our A$^2$N and other state-of-the-art lightweight networks (blue circle) on Urban100 with a scale factor of 2. }
\end{figure}

Overall, the contributions of this work are three-fold:
\begin{enumerate}
\item We quantify the effectiveness of attention layers across different stages in neural networks and propose a valid strategy for pruning attention layers.

\item We propose low-consumption attention in attention (A$^2$) structure, which can produce sum-to-one attention weights for its internal branches. 
The weights are dynamically determined by the input features.  A$^2$ structure greatly improves the capacity of the attention network with little parameters overhead. 
It can be easily applied to other methods at a very low cost, which has great potential value.
\item We propose an A$^2$ network (A$^2$N) based on A$^2$ structure, which demonstrates superior performance compared to baseline networks. Indicated by local attribution map, an SR network interpretation method, A$^2$ network also utilizes a wider range of information for better SR results.
 \cite{gu2021interpreting}.
\end{enumerate}


\section{Related Work}

\subsection{Deep CNN for SR}

Recently, CNN-based methods for single image super-resolution (SISR) have achieved significant promotion.
The  network  design  is  one  of  the  most important  parts in SR problem.
Pioneer work of SRCNN \cite{dong2015image} first introduced a shallow three-layer convolutional neural network for image SR, which shows superior performance of deep learning.
After this work, the network architecture is constantly improving.
By introducing residual learning to ease the training difficulty,
Kim \etal  proposed deeper models VDSR\cite{kim2016accurate} and DRCN \cite{kim2016deeply} with more than 16 layers.
MADNet \cite{lan2020madnet}  proposes a dense lightweight for stronger multiscale feature expression.
Lan \etal  \cite{9007041} uses novel local wider residual blocks to effectively extract the image features for SISR and proposes a cascading residual network.
EDSR \cite{lim2017enhanced} achieves significant improvement by removing unnecessary modules (batch normalization) in residual networks.
For the sake of fusing low-level and high-level features to provide richer information and details for reconstructing, 
RDN \cite{zhang2018residual}, CARN \cite{ahn2018fast}, MemNet \cite{tai2017memnet} and ESRGAN \cite{wang2018esrgan} also adopted dense connections in layer-level and block-level.
Liu \etal \cite{liu2020residual} propose a novel residual feature aggregation framework to  fully utilize the hierarchical features on the residual branches.

\begin{figure*}[!ht]
\begin{center}
   \includegraphics[width=0.95\linewidth]{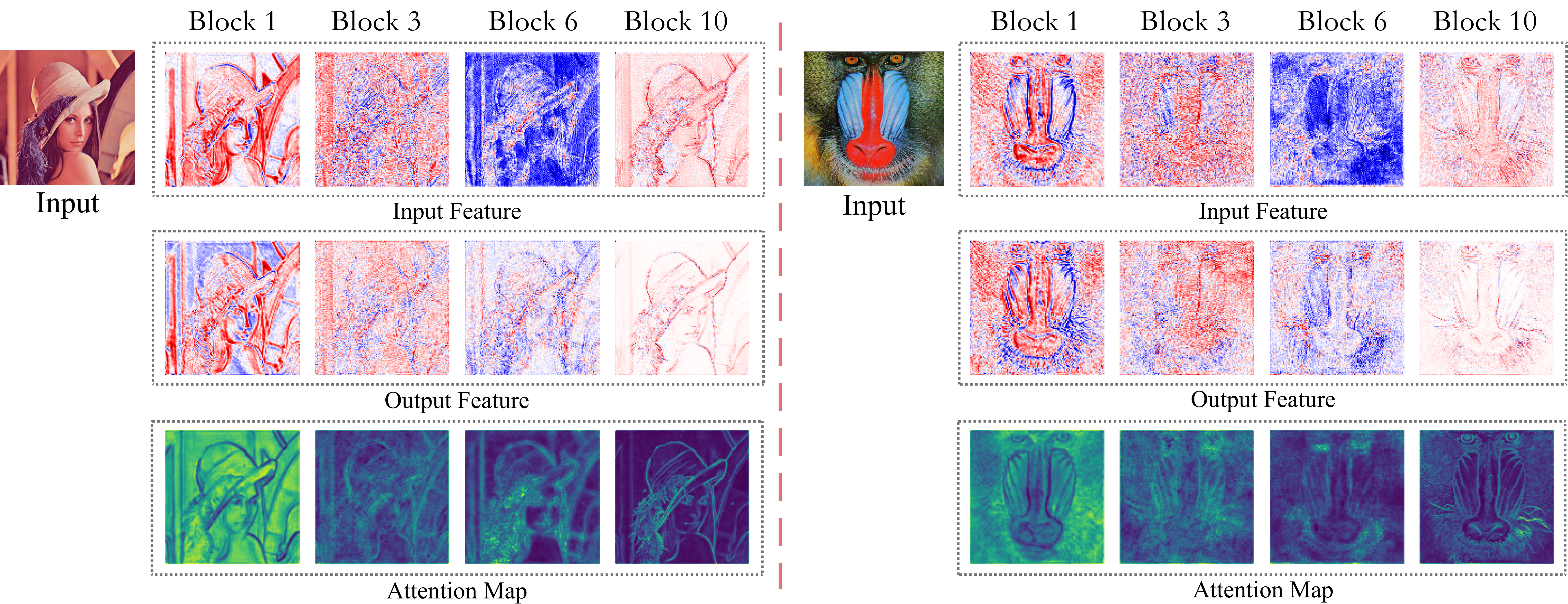}
\end{center}
   \caption{
   Attention block heatmaps.
   Due to the limited space, we chose several representative blocks, each columns indicates the first, third, sixth, tenth attention block, respectively. 
   The three rows are averaged input feature map, averaged output feature map, and the averaged attention map for each attention layer, respectively.
   For the feature maps, the white area in the feature map indicates zero values, the red area indicates positive values, and the blue area indicates negative values.
   For the attention maps, brighter colors represent a higher attention coefficient.   
}
\label{fig:f1}
\end{figure*}

\subsection{Attention Mechanism}
\label{att_related}

Attention mechanism in deep learning is similar to the attention mechanism of human vision.
It can be viewed as a means of biasing the allocation of available computational resources towards the most informative components of a signal \cite{hu2018squeeze}.
Attention mechanism usually contains a gating  function to generate a feature mask. 
It has been applied to many computer vision tasks, such as image captioning \cite{xu2015show,chen2017sca} and image classification \cite{hu2018squeeze, wang2017residual}.
Wang et al. \cite{wang2018non} initially proposed non-local operation for capturing long-range dependencies, 
it computes the response at a position as a weighted sum of the features at all positions.
Hu  et  al.  \cite{hu2018squeeze} focus on the channel relationship and propose a novel architectural unit, Squeeze-and-Excitation (SE) block, that adaptively recalibrates channel-wise feature responses. 
Woo et  al. \cite{woo2018cbam}  provided a study on the combination of channel and spatial attention.

In recent years, several works proposed to investigate the effect of attention mechanism on low-level vision tasks.
\cite{liu2018non} first attempt to incorporate non-local operations into a recurrent neural network for image restoration.
RNAN \cite{zhang2019residual} proposed residual local and non-local attention blocks in the mask branch in order to obtain non-local mixed attention.
Channel attention is another popular way to embed attention mechanism.
RCAN \cite{zhang2018image}  exploits the interdependencies among feature channels by generating different attention for each channel-wise feature.
Some works utilize both channel attention and non-local attention.
SAN \cite{dai2019second} performed region-level non-local operations for reducing computational burden,
and proposed second-order channel attention by considering second-order statistics of features. 
%


\section{Motivation}

Given an intermediate feature map $\mathbf{F} \in \mathbb{R}^{C\times H\times W}$, where $C$, $H$, and $W$ are the number of channels, height, and width of the features, respectively. 
Attention mechanism infers an attention map function $\mathbf{M}_{\mathbf{A}}$, where $\mathbf{M}_{\mathbf{A}}(\mathbf{F}) \in \mathbb{R}^{C^{\prime} \times H^{\prime} \times W^{\prime}}$, 
the size of $C^{\prime} $, $ H^{\prime} $ and $ W^{\prime}$ depend on the type of attention function.
For example, channel attention generates a 1D ($\mathbb{R}^{C \times 1 \times 1}$) channel-wise attention vector \cite{hu2018squeeze,zhang2018image,woo2018cbam,park2018bam,hu2019channel}. 
Spatial attention generates a 2D ($\mathbb{R}^{1 \times H \times W}$) attention mask,
\cite{woo2018cbam,park2018bam,hu2019channel}.
Channel-spatial attention generates 3D ($\mathbb{R}^{C \times H \times W}$) attention map,
\cite{zhang2019residual,zhao2020efficient}.
We naturally raise two questions: 
(1) Which part of an image tends to have a higher or lower attention coefficient? 
(2) Are attention mechanisms always beneficial to SR models? 

\subsection{Attention Heatmap}

\begin{table*}[]
\renewcommand\arraystretch{1.2}
\centering
\caption{The correlation coefficients between the attention map and the output feature of high-pass filters of the corresponding feature map for each attention block. {\color[HTML]{FE0000} Red}/{\color[HTML]{3531FF} Blue} text: maximum and minimum value. The bottom two rows are the mean and standard deviation of each attention map.}
\label{tab:corr}
\begin{tabular}{lcccccccccc}
\toprule
\multirow{2}{*}{Filter} & \multicolumn{10}{c}{Correlation Coefficients with the Input Feature}                                                      \\ \cmidrule{2-11} 
                        & Block 1 & Block 2 & Block 3 & Block 4 & Block 5 & Block 6 & Block 7 & Block 8 & Block 9 & Block 10 \\ \midrule
Laplace Operator        & {\color[HTML]{3531FF} -0.270}  & -0.082  & 0.323   & 0.281   & 0.296   & 0.158   & -0.098  & 0.100   & 0.184   & {\color[HTML]{FE0000} 0.433}    \\
Scharr Operator         & {\color[HTML]{3531FF} -0.395}  & -0.122  & 0.398   & 0.355   & 0.412   & 0.206   & -0.079  & 0.073   & 0.255   & {\color[HTML]{FE0000} 0.435}    \\
Sobel Operator          & {\color[HTML]{3531FF} -0.393}  & -0.153  & 0.371   & 0.334   & 0.379   & 0.175   & -0.109  & 0.090   & 0.196   & {\color[HTML]{FE0000} 0.491}    \\ \midrule \midrule 
Mean of Attention Map         & 0.208  & 0.126  & 0.078 & 0.106 & 0.065 & 0.035 & 0.048  & 0.039 & 0.051 & 0.086 \\ 
Std of Attention Map         & 0.125  & 0.070  & 0.042 & 0.081 & 0.055 & 0.039 & 0.060  & 0.039 & 0.027 & 0.075 \\ \bottomrule
\end{tabular}
\end{table*}

\begin{figure}[!t]
\begin{center}
   \includegraphics[width=0.92\linewidth]{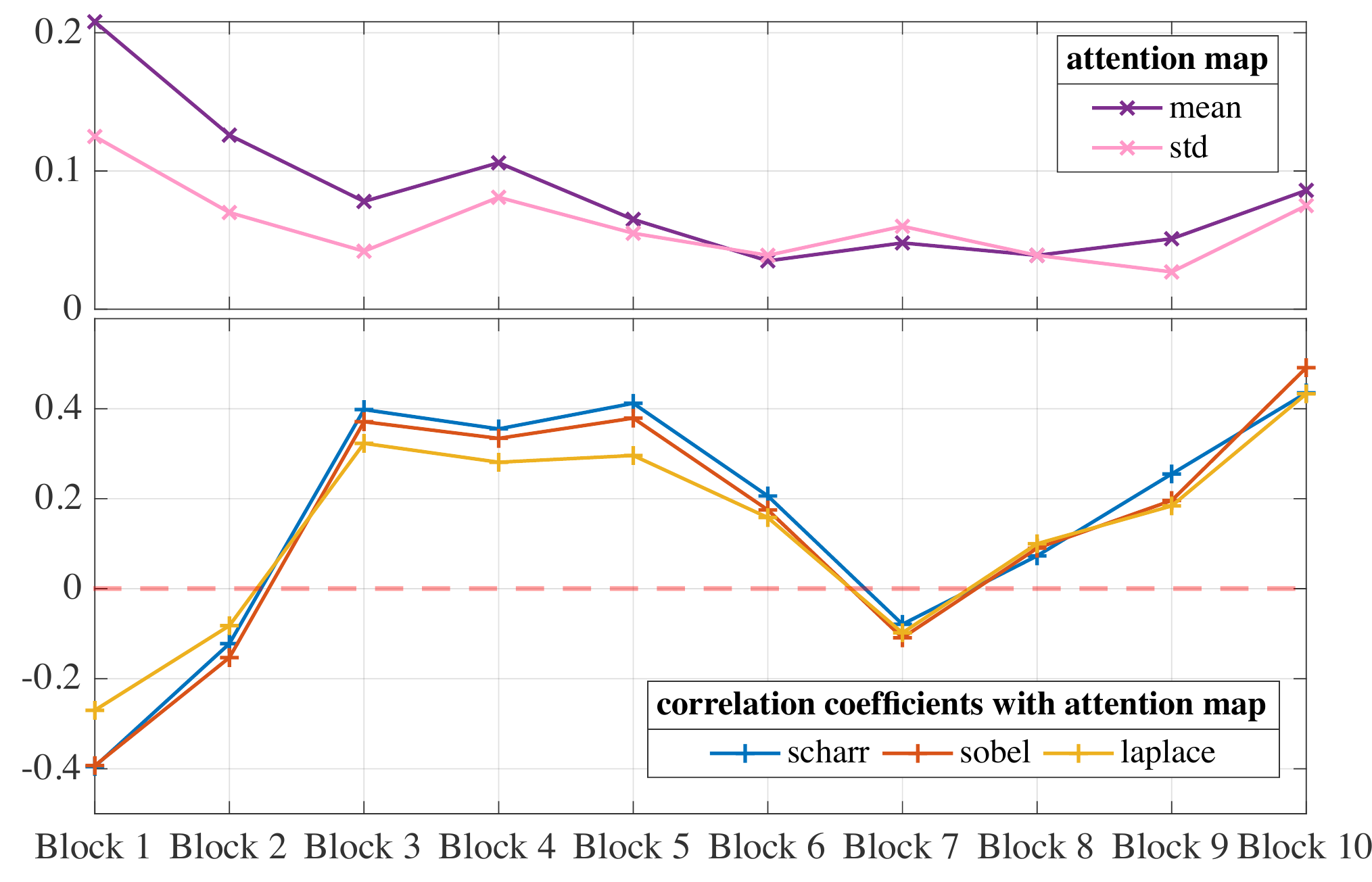}
\end{center}
   \caption{The correlation coefficients between the attention map and the output feature of high-pass filters of the corresponding feature map for each attention block.}
\label{fig:net}
\end{figure}

\begin{table}[!htbp]
\renewcommand\arraystretch{1.1}
\begin{center}
\caption{Attention layer importance measurement. The first column indicates the insertion stage of the residual attention blocks.}
\label{tab:loc}
\begin{tabular}{ccc}
\toprule
Attention Block Index           & \# Parameter          & PSNR  \\ \midrule
All           & 9.2 M           & \textbf{28.65}      \\ 
None          & 4.4 M           & 28.60       \\
\{1,2,3,4,5\}        & 6.8 M    & 28.60         \\ 
\{6,7,8,9,10\}       & 6.8 M    & \textbf{28.65}     \\ 
\{2,4,6,8,10\}       & 6.8 M    & 28.63           \\ \bottomrule
\end{tabular}
\end{center}
\end{table}

Previous work \cite{zhang2018image} suggests that information in the LR space has abundant low-frequency and valuable high-frequency components, 
all features are treated equally without using the attention mechanism in the networks, while attention can help the network pay more attention to the high-frequency features.
The high-frequency components would usually be regions, being full of edges, texture, and other details \cite{zhang2018image}.
However, to our best knowledge, few works truly prove the above assumptions.
To answer the first question, we conduct experiments to understand the behavior of the attention mechanism in SR. 
We construct a simple attention-based model, which consists of ten attention blocks.
Each attention block uses a channel- and spatial-wise attention layer so that every pixel has an individual attention coefficient.
We use the sigmoid function as the gating function so that the attention coefficient can be scaled into $[0, 1]$. 
We visualize some feature maps and attention maps in \figurename~\ref{fig:f1}. \tablename~\ref{tab:corr} lists correlation coefficients between the attention map and the high-pass filtering results to the corresponding feature map.
Note that this is not a highly accurate method to measure the exact attention response, but our intention is to quantify the relative high-pass correlation across different layers.
Based on the observations from the visualized attention maps in \figurename~\ref{fig:f1} and \tablename~\ref{tab:corr}, we show that attentions learnt at different layers vary a lot with respect to their relative depth in the neural network.
For example, the first and tenth attention blocks show opposite responses.
The attention modules in the first two layers tend to enhance the low-frequency bands of the features, while the last two layers enhance the high-frequency features.
The blocks in-between have mixed responses.
%

\subsection{Ablating Attention}
\label{att_dropout}

Based on the above results,  we may be able to maximize the use of attention while minimizing the number of additional parameters. 
An intuitive idea is to preserve attention layers only at performance-critical layers. However, the above qualitative analysis is not a valid method to quantify the realistic effect of attention layers.
To quantitatively measure the effectiveness of attention layers, we propose a dynamic attention framework. 
An attention block, regardless of its type mentioned previously in Section~\ref{att_related}, can be downgraded to a non-attentional block by simply removing the attention generator operation.

We have conducted a series of experiments with certain attention layers turned off.
The results are shown in \tablename~\ref{tab:loc}, where the first column indicates residual attention blocks that are enabled. For example, \{1, 2, 3, 4, 5\} means attention layers in the first five blocks are residual attention blocks while others are turned off and downgraded to basic residual blocks.
The results lead to an interesting result: the relative block depth matters a lot in the decision of where to insert attention blocks. Enabling \{6, 7, 8, 9, 10\} blocks with attention is effectively achieving the same PSNR as tuning on every block but with much fewer parameters.
This experiment further proves that spending budget on attention uniformly across the network is sub-optimal.

\begin{figure*}[!t]
\begin{center}
   \includegraphics[width=0.85\linewidth]{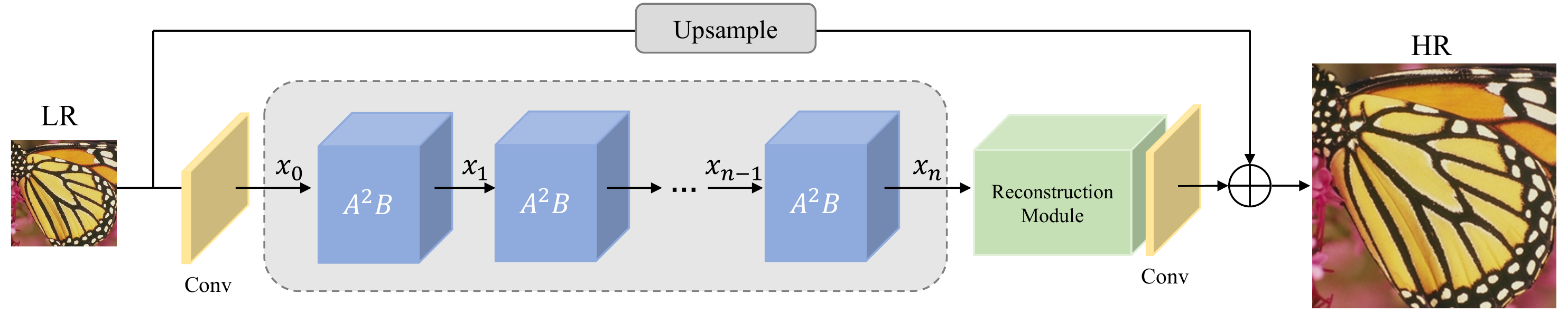}
\end{center}
   \caption{Overview of attention in attention network (A$^2$N).}
\label{fig:net}
\end{figure*}

\begin{figure}[!h]
\begin{center}
   \includegraphics[width=0.85\linewidth]{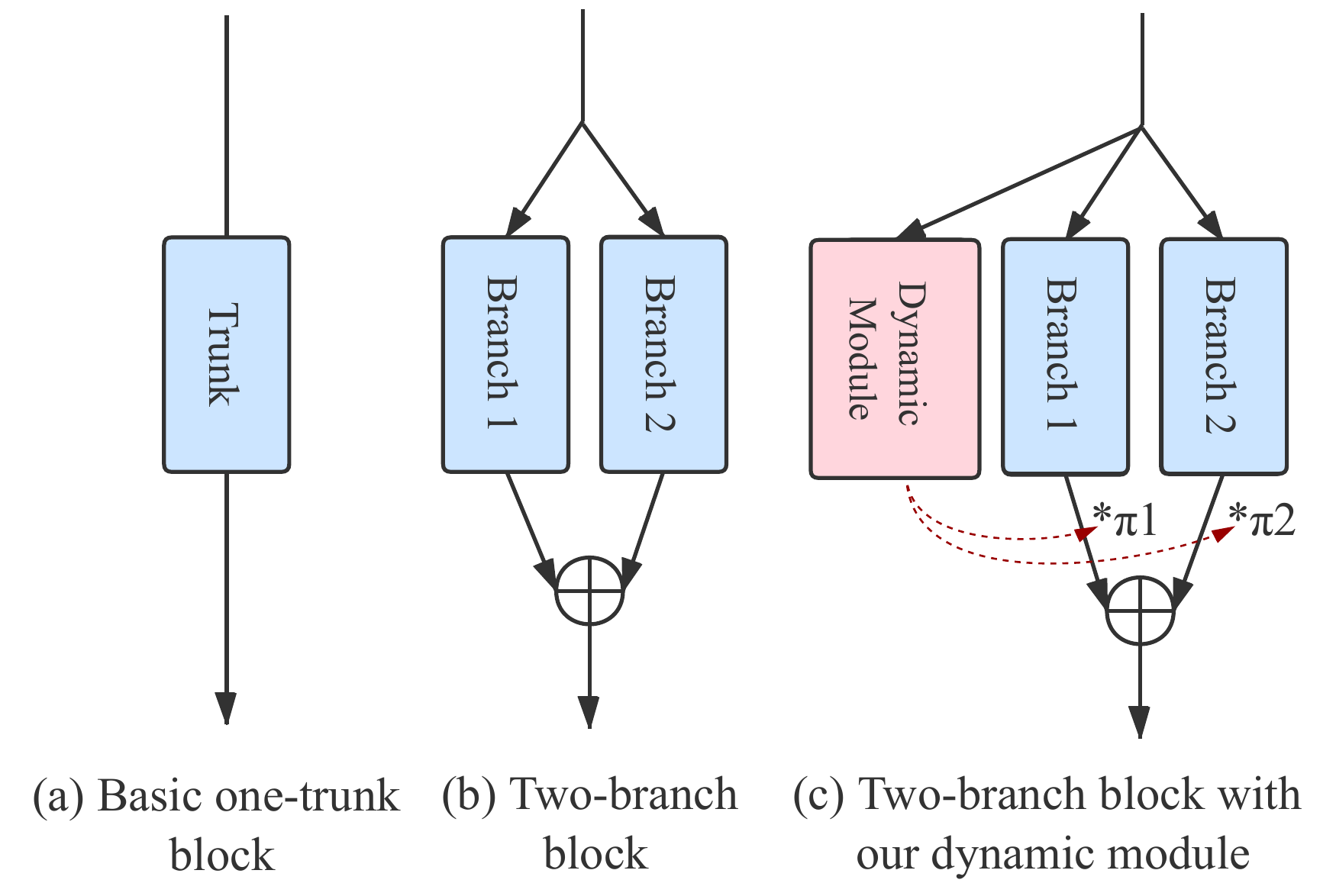}
\end{center}
   \caption{Dynamic Attention Module.}
\label{fig:comparison}
\end{figure}


\section{Method}

Previous models \cite{zhang2018image,zhao2020efficient,hu2019channel} with fixed attention layers have attention maps activated all the time, regardless of image content. We have shown in Sec.~\ref{att_dropout} that the effectiveness of attention layers varies at different locations in the neural network. Our motivation here is to create non-attentional shortcut branches for counter-part attention branches with mixing weights generated dynamically through an additional evaluation module using the same input features as ordinary layers. 

\subsection{Network Architecture}

As shown in Figure~\ref{fig:net}, the network architecture of our proposed method, consists of three parts: shallow feature extraction,  attention in attention block deep feature extraction, and reconstruction module. The input and output image is denoted as $I_{LR}$ and $I_{SR}$. Follow \cite{lim2017enhanced}, we use a single convolution layer in shallow feature extraction module. We can then formulate
$x_{0}=f_{ext}(I_{LR})$
, where $f_{ext}(\cdot)$ is a convolution layer with 3$\times$3 kernel size to extract the shallow feature from the input LR image $I_{LR}$, $x_0$ is the extracted feature map.

We construct our deep feature extractor as a chained sub-network using A$^2$B.
\begin{equation}
x_{n}=f_{A^2 B}^{n}\left(f_{A^2 B}^{n-1}\left(\ldots f_{A^2 B}^{0}\left(x_{0}\right) \ldots\right)\right)
\end{equation}
where $f_{A^2 B}(\cdot)$ denotes the attention in attention block. A$^2$B combines non-attention branch and attention branch with dynamic weights determined by the input feature.

After deep feature extraction, we upscale the deep feature $x_n$ via the reconstruction module.
In the reconstruction module, we first use nearest-neighbor interpolation for upsampling, 
then we use a simplified channel-spatial attention layer between two convolution layers. 
This simplified attention layer only uses one 1 $\times$ 1 convolution and sigmoid function to generate the attention map.
We also use global connection, in which a nearest-neighbor interpolation is performed on the input $I_{LR}$.
The final model produces high resolution result by applying the reconstruction signal to upsampled output: 
\begin{equation}
I_{SR} = f_{rec}(x_n) + f_{up}(I_{LR})
\end{equation}
$f_{rec}(\cdot)$ is the reconstruction module, $f_{up}(\cdot)$ is the bilinear interpolation.  $I_{SR}$ is the final SR output.
    
\begin{figure*}[!t]
\begin{center}
   \includegraphics[width=0.85\linewidth]{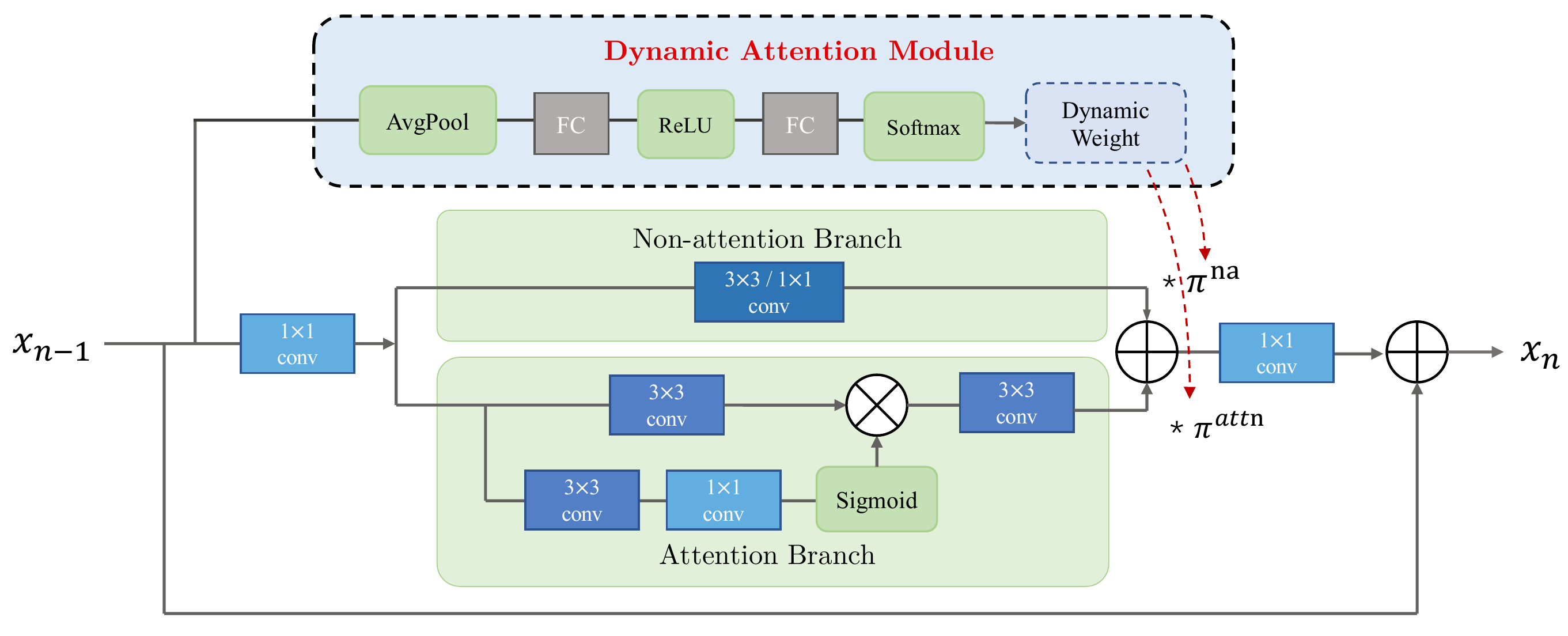}
\end{center}
   \caption{Architecture of the attention in attention block (A$^2$B).}
\label{fig:dab}
\end{figure*}

\subsection{Attention in Attention Block (A$^2$B)}

We have discussed and dynamic contribution from different attention layers in section~\ref{att_dropout}, 
nevertheless, it is infeasible to manually determine the topological structure of attention modules.
Inspired by the dynamic kernel \cite{chen2020dynamic} which use dynamic convolution to aggregate multiple parallel convolution kernels dynamically based upon their attentions, 
here we propose a learnable dynamic attention module to automatically drop some unimportant attention features and balance the attention branch and non-attention branch.
More specifically, each dynamic attention module controls the dynamic weighted contribution from the attention and the non-attention branch using weighted summation. As depicted in Figure~\ref{fig:comparison}.
Dynamic attention module generates weights by using the same input feature of its block as two independent branches. 
Formally, we have:

\begin{equation}
\label{eq:dwm}
    x_{n+1} =  f_{1\times1}(\pi_{n}^{na} \times x_{n}^{na} + \pi_{n}^{attn} \times x_{n}^{attn})
\end{equation}
where $x_{n}^{na}$ is the output of non-attention branch, and $x_{n}^{attn}$ is the output of the attention branch.
$f_{1\times 1}(\cdot)$ donates 1 $\times$ 1 kernel convolution.
$\pi_{na}$ and $\pi_{attn}$ are weights of non-attention branch and attention branch respectively, 
they are computed by the network according to the input feature, instead of two fixed values which are artificially set.

To compute the dynamic weights, we have:
\begin{equation}
    \pi_{n} = f_{da}(x_{n})
\end{equation}
where $f_{da}(\cdot)$ is the dynamic attention module. 
The dynamic attention module can be viewed in detail in Figure~\ref{fig:dab}. It firstly squeezes the input $x_{n-1}$ using global average pooling. The connecting layers consist of two fully connected layers with a ReLU activation.
We use global pooling to increase receptive field, which allows dynamic attention module to capture features from the whole image, section~\ref{local_attribution_maps} gives experiments to prove this.
It is worth mentioning that the dynamic attention module is also used in the inference stage, 
once the input feature changes, the weights of the two branches  also change.

As investigated in \cite{chen2020dynamic}, constraining the dynamic weights can facilitate the learning of the dynamic attention modules.  Specifically, we have the sum-to-one constraint $\pi_n^{na} + \pi_n^{attn} = 1$. This sum-to-one constraint for the dynamic weights can compress the kernel space. It significantly simplifies the learning of $\pi$. 
Therefore, a softmax function is followed to generate normalized attention weights for the two branches.

The overall structure of our proposed attention in attention block in shown in Figure~\ref{fig:dab}. $\bigotimes$ combines feature and attention map by element-wise multiplication, $\bigoplus$ computes weighted summation over two branches as Eq.~\ref{eq:dwm}.


\begin{figure}[!htbp]
\begin{center}
   \includegraphics[width=0.99\linewidth]{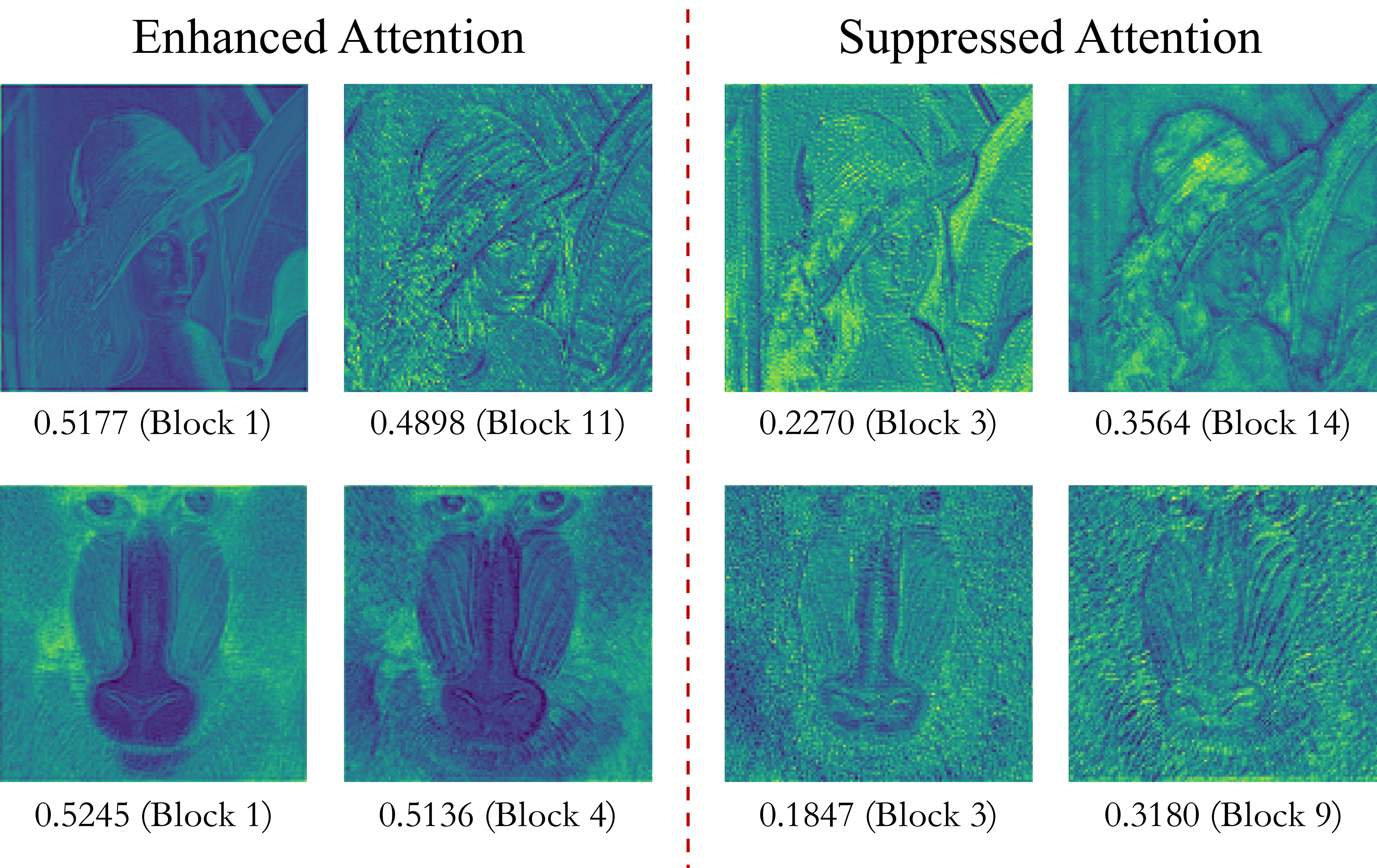}
\end{center}
\caption{\textbf{left} The most enhanced attention maps. \textbf{right} The most suppressed attention maps.
The weight value $\pi^{attn}$ and the block index are shown under the attention maps.
}
\label{fig:ab2}
\label{fig:visual}
\end{figure}

\section{Experiments}

\begin{table*}[]
\renewcommand\arraystretch{1.3}
\centering
\caption{The effect of A$^2$ structure. Test on Set14 ($\times$ 4).}
\label{tab:ab}
\begin{tabular}{clccccccccc}
\toprule\vspace{2pt}
 &
   &
  \multicolumn{2}{c}{Branch} &
  \multicolumn{5}{c}{Branch Fusion} &
  \multicolumn{2}{c}{Performance} \\ \toprule
 &
   &
  \cellcolor[HTML]{EFEFEF}\rotatebox{90}{Non-attention} &
  \cellcolor[HTML]{EFEFEF}\rotatebox{90}{Attention} &
  \rotatebox{90}{Addition} &
  \rotatebox{90}{Concatenation} &
  \rotatebox{90}{Adaptive-Weight} &
  \rotatebox{90}{A$^2$ (1$\times$1 conv)} &
  \rotatebox{90}{A$^2$ (3$\times$3 conv)} &
  \cellcolor[HTML]{F1F1F1}\rotatebox{90}{Parameter (K)} &
  \cellcolor[HTML]{F1F1F1}\rotatebox{90}{PSNR} \\ \midrule
 &
  A$^2$N-non-attn-only &
  \cellcolor[HTML]{EFEFEF}$\bullet$ &
  \cellcolor[HTML]{EFEFEF} &
   &
   &
   &
   &
   &
  \cellcolor[HTML]{F1F1F1}208 &
  \cellcolor[HTML]{F1F1F1}28.515 \\
\multirow{-2}{*}{Single branch} &
  A$^2$N-attn-only (baseline) &
  \cellcolor[HTML]{EFEFEF} &
  \cellcolor[HTML]{EFEFEF}$\bullet$ &
   &
   &
   &
   &
   &
  \cellcolor[HTML]{F1F1F1}810 &
  \cellcolor[HTML]{F1F1F1}28.646 \\ \midrule
 &
  A$^2$N-Addition &
  \cellcolor[HTML]{EFEFEF}$\bullet$ &
  \cellcolor[HTML]{EFEFEF}$\bullet$ &
  $\bullet$ &
   &
   &
   &
   &
  \cellcolor[HTML]{F1F1F1}1,040 &
  \cellcolor[HTML]{F1F1F1}28.651 \\
 &
  A$^2$N-Concatenation &
  \cellcolor[HTML]{EFEFEF}$\bullet$ &
  \cellcolor[HTML]{EFEFEF}$\bullet$ &
   &
  $\bullet$ &
   &
   &
   &
  \cellcolor[HTML]{F1F1F1}1,092 &
  \cellcolor[HTML]{F1F1F1}28.642 \\
\multirow{-3}{*}{\begin{tabular}[c]{@{}c@{}}Two Branches,\\ without A$^2$\end{tabular}} &
  A$^2$N-Adaptive-Weights &
  \cellcolor[HTML]{EFEFEF}$\bullet$ &
  \cellcolor[HTML]{EFEFEF}$\bullet$ &
   &
   &
  $\bullet$ &
   &
   &
  \cellcolor[HTML]{F1F1F1}1,040 &
  \cellcolor[HTML]{F1F1F1}28.648 \\ \midrule
 &
  A$^2$N-S (Fewer Channels ) &
  \cellcolor[HTML]{EFEFEF}$\bullet$ &
  \cellcolor[HTML]{EFEFEF}$\bullet$ &
   &
   &
   &
   &
  $\bullet$ &
  \cellcolor[HTML]{F1F1F1}678 &
  \cellcolor[HTML]{F1F1F1}28.651 \\
 &
  A$^2$N-M &
  \cellcolor[HTML]{EFEFEF}$\bullet$ &
  \cellcolor[HTML]{EFEFEF}$\bullet$ &
   &
   &
   &
  $\bullet$ &
   &
  \cellcolor[HTML]{F1F1F1}843 &
  \cellcolor[HTML]{F1F1F1}28.695 \\
\multirow{-3}{*}{A$^2$} &
  A$^2$N &
  \cellcolor[HTML]{EFEFEF}$\bullet$ &
  \cellcolor[HTML]{EFEFEF}$\bullet$ &
   &
   &
   &
   &
  $\bullet$ &
  \cellcolor[HTML]{F1F1F1}1,047 &
  \cellcolor[HTML]{F1F1F1}28.707 \\ \bottomrule
\end{tabular}
\end{table*}

In this section, we compare our method with state-of-the-art SISR
algorithms on five commonly used benchmark datasets. 
Besides, we conduct ablation study to validate and analyze the effectiveness of our proposed method.

\subsection{Datasets and Metrics}

We use DIV2K dataset \cite{Agustsson_2017_CVPR_Workshops} as our training dataset, which contains 800 training images. 
The LR images are obtained by the bicubic downsampling of HR images. 
For testing stage, we use five standard benchmark datasets: Set5\cite{bevilacqua2012low}, Set14\cite{yang2010image}, B100\cite{martin2001database}, Urban100\cite{huang2015single} and Manga109\cite{matsui2017sketch}. 
The SR results are evaluated by peak signal to noise ratio (PSNR) and the structural similarity index (SSIM) on the Y channel of YCbCr space.

\subsection{Implementation Details}
Now we specify the implementation details of our proposed A$^2$N. 
We design two variants of A$^2$N, denoted as A$^2$N and A$^2$N-M.
On the non-attention branch, we use 3$\times$3 convolution for A$^2$N and 1$\times$1 convolution for A$^2$N-M.
For both variants, we set the number of A$^2$B as 16. 
Features in A$^2$B have 40 filters, except for that in the upsampling block, where C = 24.
For the training process,
data augmentation is performed on the 800 training images, which are randomly rotated by 90\degree, 180\degree, 270\degree, and flipped horizontally. 

\subsection{Results}
\label{experiment_1}

\begin{table*}[]
\renewcommand\arraystretch{1.3}
\caption{Ablation study: effect of different components. Test on Set14 ($\times$ 4).}
\centering
\label{tab:components}
\begin{tabular}{ccccccccc}
\toprule
Case Index                & \cellcolor[HTML]{F3F3F3} 1          & \cellcolor[HTML]{F3F3F3} 2          & \cellcolor[HTML]{F3F3F3} 3          & \cellcolor[HTML]{F3F3F3} 4          &  5          &  6          &  7          &  8          \\ \midrule
non-attention             & \cellcolor[HTML]{F3F3F3} $\bullet$  & \cellcolor[HTML]{F3F3F3}            & \cellcolor[HTML]{F3F3F3} $\bullet$  & \cellcolor[HTML]{F3F3F3} $\bullet$  &  $\bullet$  &             &  $\bullet$  &  $\bullet$ \\
channel attention         & \cellcolor[HTML]{F3F3F3} $\bullet$  & \cellcolor[HTML]{F3F3F3} $\bullet$  & \cellcolor[HTML]{F3F3F3}            & \cellcolor[HTML]{F3F3F3}            &  $\bullet$  &  $\bullet$  &             &              \\
spatial attention         & \cellcolor[HTML]{F3F3F3}            & \cellcolor[HTML]{F3F3F3} $\bullet$  & \cellcolor[HTML]{F3F3F3}            & \cellcolor[HTML]{F3F3F3} $\bullet$  &             &  $\bullet$  &             &  $\bullet$ \\
channel-spatial attention & \cellcolor[HTML]{F3F3F3}            & \cellcolor[HTML]{F3F3F3}            & \cellcolor[HTML]{F3F3F3} $\bullet$  & \cellcolor[HTML]{F3F3F3}            &             &             &  $\bullet$  &             \\ 
\textbf{A$^2$}                     & \cellcolor[HTML]{F3F3F3}            & \cellcolor[HTML]{F3F3F3}            & \cellcolor[HTML]{F3F3F3}            & \cellcolor[HTML]{F3F3F3}            &  $\bullet$  &  $\bullet$  &  $\bullet$  &  $\bullet$ \\ \midrule
Parameter (K)             & \cellcolor[HTML]{F3F3F3} 787        & \cellcolor[HTML]{F3F3F3} 1,035      & \cellcolor[HTML]{F3F3F3} 1,040      & \cellcolor[HTML]{F3F3F3} 791        &  794        &  1,042      &  1,047      &  798       \\
PSNR                      & \cellcolor[HTML]{F3F3F3} 28.634     & \cellcolor[HTML]{F3F3F3} 28.649     & \cellcolor[HTML]{F3F3F3} 28.651     & \cellcolor[HTML]{F3F3F3} 28.583     &  28.600     &  28.629     &  28,707     &  28.642     \\
Gain from A$^2$           & \cellcolor[HTML]{F3F3F3} -          & \cellcolor[HTML]{F3F3F3} -          & \cellcolor[HTML]{F3F3F3} -          & \cellcolor[HTML]{F3F3F3} -          &  -0.034    &  -0.020     &  +0.056    &  +0.059     \\ \bottomrule  
\end{tabular}
\end{table*}

To demonstrate the effect of our proposed A$^2$ structure, 
we compare our two-branch A$^2$ structure with a one-trunk structure.
The results are shown in \tablename~\ref{tab:ab}.
If we only keep the attention branch, similar to previous models, 28.646 dB is obtained with 810K parameters.
Using our A$^2$ structure, the performance increased by 0.05 dB with only 200K extra parameters; 
If we reduce the number of channels of our method to 32, A$^2$ performs better with 132K fewer parameters.
The results also prove once again that not all attention layers are making positive contributions.

We also show results in \tablename~\ref{tab:components} to evaluate the performance of A$^2$ structure on the multi-branch model.
Cases 1-4 are two-branch models without A$^2$ structure, features from two branches are fused by an addition operation.
Cases 5-8 modify cases 1-4 by applying A$^2$ structure.
As we can see, for cases 1-4, the combination of non-attention and channel-spatial attention has the best performance. Therefore, we use channel-spatial attention in the attention branch.
For cases 1-4, the non-attention branch combine with spatial attention or channel-spatial attention (cases 6 and 7) gains more than 0.05 dB by only about 7K parameters cost.
Therefore, the dynamic attention module performs well when used in models without pooling or downsampling layers.

We compare our method with various SR methods of similar model sizes: 
SRCNN \cite{dong2015image}, FSRCNN \cite{dong2016accelerating}, DRRN \cite{tai2017image}, VDSR \cite{kim2016accurate}, MemNet \cite{tai2017memnet}, IMDN \cite{hui2019lightweight}, A$^2$F-M \cite{wang2020lightweight}, AWSRN-M \cite{wang2019lightweight}, SRMDNF \cite{zhang2018learning}, CARN \cite{ahn2018fast},  MADNet\cite{lan2020madnet} and DRCN \cite{kim2016deeply}.
We also selected 3 large networks for reference, ERN \cite{9007041}, RCAN \cite{zhang2018image}, and EDSR \cite{lim2017enhanced}.
\tablename~\ref{tab:sota}  shows quantitative comparisons for $\times$2, $\times$3, and $\times$4 SR. 
Note that we only compare models which have a similar number of parameters to our models in this table.
Our A$^2$N can achieve comparable or better results than state-of-the-art methods for all scaling factors. 
In particular, A$^2$N-M, which has about 200K fewer parameters than A$^2$F-M and AWSRN-M, achieves better results than these models on most datasets.
For Manga109 ($\times 3$), the PSNR of A$^2$N-M is 0.15 dB higher than the PSNR of AWSRN-M.
For more difficult scenarios, such as scales 3 and 4, the advantages of our method are more obvious. The performance on more challenging datasets such as Urban100 and Manga109 is more than 0.2 dB higher than that of IMDN.

It should be noted that our core contribution is to dynamically adjust attention to make it play a greater potential, not in terms of lightweight. Therefore, we have not made improvements in terms of lightweight. And in order to better prove the effectiveness of our Dynamic Attention Module, we also did not make a special design for the attention branch, but used the simplest channel-spatial attention.
We prove the effectiveness of the Dynamic Attention Module in sections 1 and 2 below, and prove its effect on large models in section 3.

Figure~\ref{fig:visual} shows the visual comparison for upscaling factor $\times$4.
We can see that our method achieves better performance than others, it can recover high frequency details more accurately.
For image “zebra”, our method can restore zebra stripes accurately.
For the details of the buildings, in image "78004", the windows are restored more clearly and accurately.
For image "img092", only our method can restore the direction of the fringe correctly.

\subsection{Discussion}

We use the weight prediction in the dynamic attention layer to sample the enhanced and suppressed attention signals.
Figure~\ref{fig:ab2} shows the two attention maps with the top two weights and the two attention maps with the smallest weights.
For each row, 
the attention maps selected by the highest and second-highest weights are listed on the left, while
the right depicts the two attention maps with the smallest weights.
As we can see, for attention maps with high weights, the edge and texture components of the feature map can be accurately located,
while the attention maps with lower weights can not found it accurately. 
These results demonstrate the effectiveness of the dynamic attention module, which can automatically determine the weights of the two branches based on the input features.

\begin{figure}[!htbp]
\begin{center}
   \includegraphics[width=0.95\linewidth]{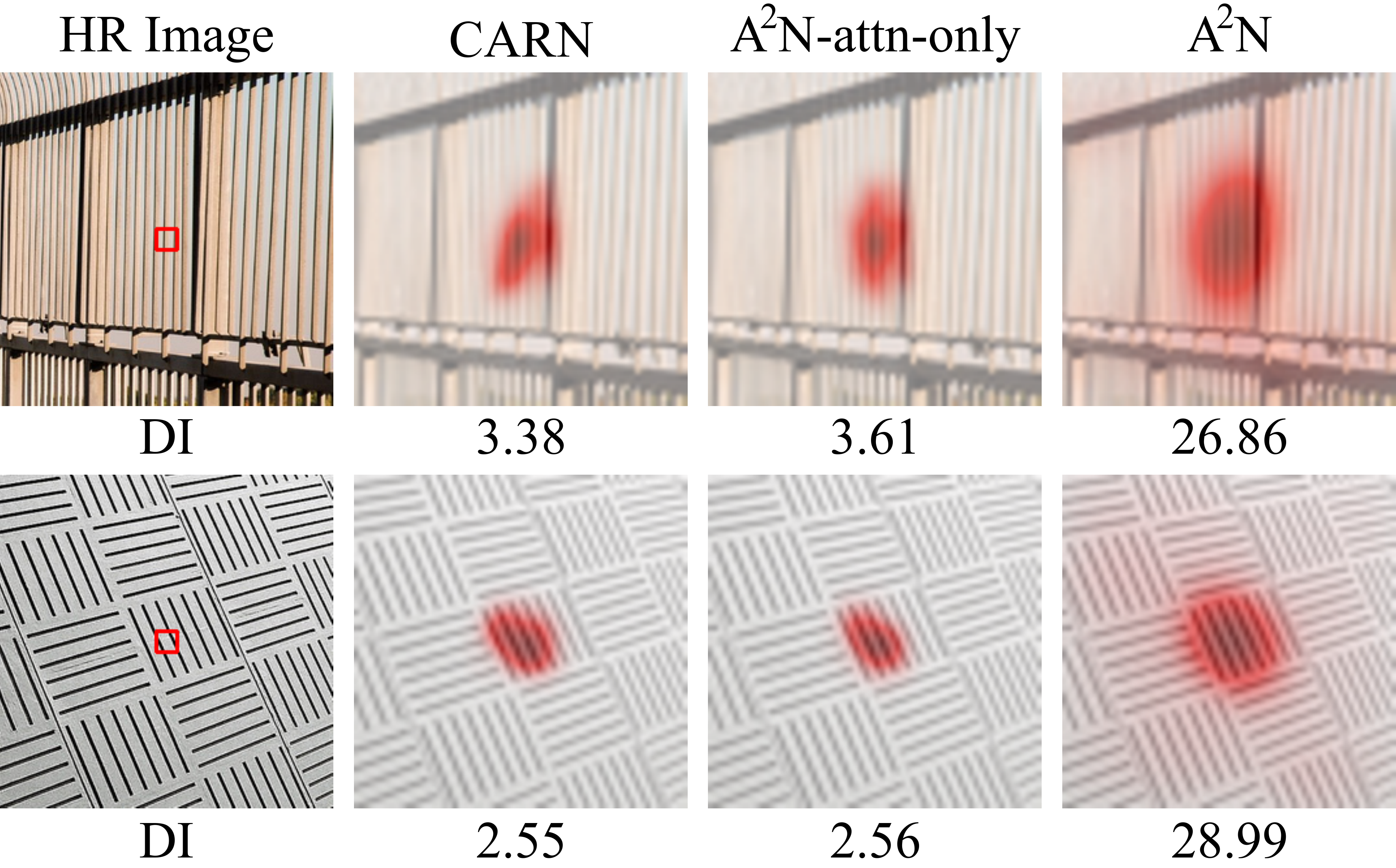}
\end{center}
\caption{Results of the LAM for SR network interpretation. The LAM maps represent the importance of each pixel in the input LR image w.r.t. the SR of the patch marked with a red box. We illustrate the area of contribution in red color.}
\label{fig:lam}
\end{figure}

\begin{figure*}[!t]
\begin{center}
   \includegraphics[width=0.9\linewidth]{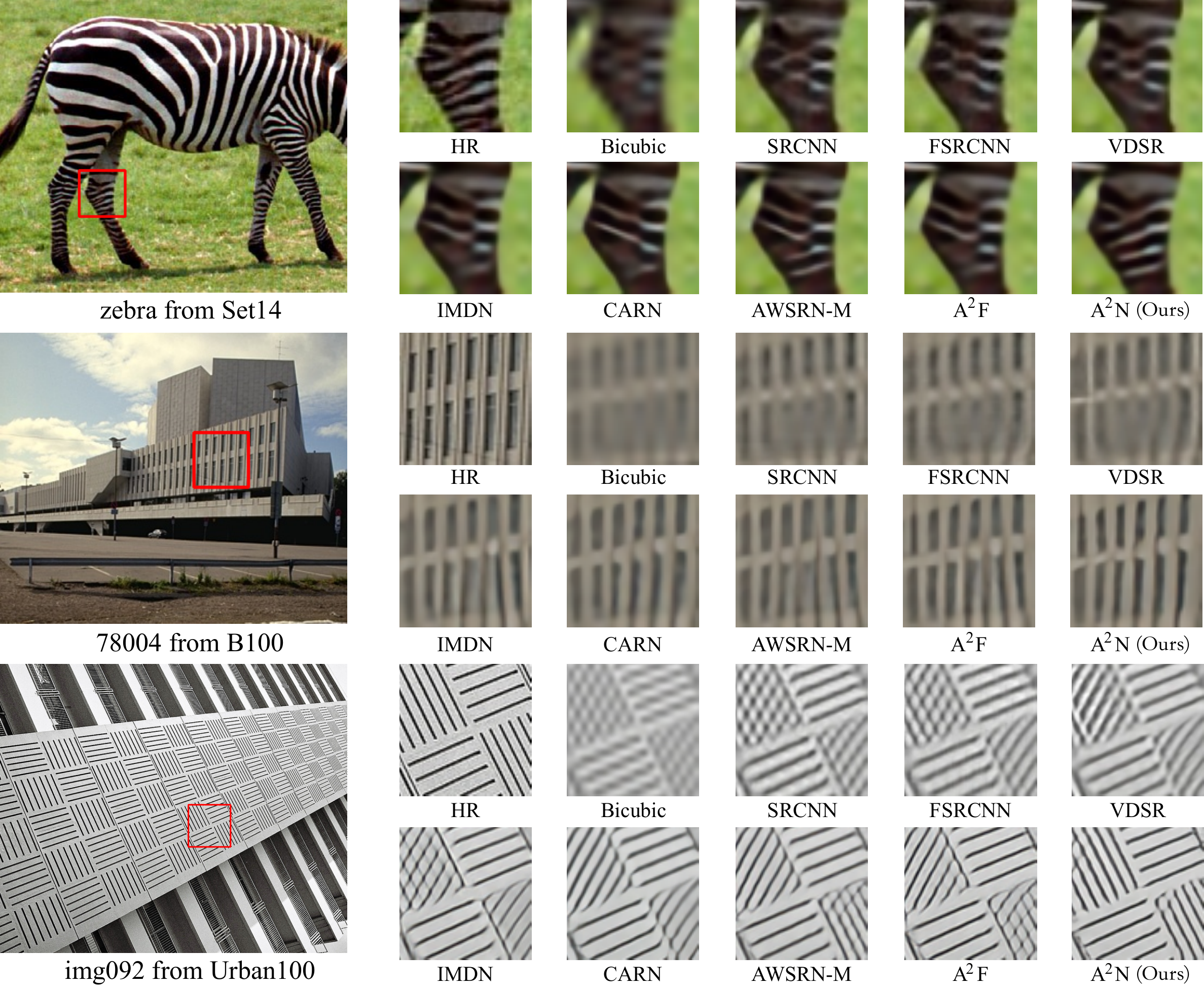}
\end{center}
   \caption{Visual comparison for upscaling factor $\times$4.}
\label{fig:dab}
\end{figure*}

\begin{table}[]
\renewcommand\arraystretch{1.2}

\begin{center}

\caption{Results of different methods used in non-attention branch. A$^2$N-attn-only: no non-attention branch. 
A$^2$N-no-op: pass through in non-attention branch. Test on Set14 ($\times$ 4).}
\label{tab:non-att}
\begin{tabular}{lccc}
\toprule
Method               & Operation          & Parameter & PSNR   \\ \midrule
A$^2$N-attn-only     &  -                & 810K    & 28.646 \\ 
A$^2$N-no-op  & -                     & 817K    & 28.660 \\
A$^2$N-M      & 1 $\times$ 1 Conv   & 843K    & 28.695 \\
A$^2$N        & 3 $\times$ 3 Conv     & 1047K   & 28.707 \\ \bottomrule
\end{tabular}
\end{center}
\end{table}

\begin{table}[]
\renewcommand\arraystretch{1.2}
\centering
\caption{Test on 150 images which are proposed from LAM \cite{gu2021interpreting}. The DI reflects the range of involved pixels. A higher DI represents a wider range of attention.}
\label{tab:lam}
\begin{tabular}{lcc}
\toprule
Model           & DI     & PSNR  \\ \midrule
FSRCNN          & 0.797  & 20.30 \\
CARN            & 1.807  & 21.27 \\
IMDN            & 14.643 & 21.23 \\
A$^2$F             & 12.58  & 21.43 \\ \midrule
A$^2$N-attn-only        & 2.54   & 21.33 \\
A$^2$N-non-attn-only       & 1.56   & 20.99 \\ \midrule
A$^2$N-Addition               & 2.75   & 21.37 \\
A$^2$N-Concatenation          & 2.70  & 21.38 \\
A$^2$N-AdaptiveWeights       & 2.43   & 21.33 \\ \midrule
\textbf{A$^2$N}             & \textbf{14.77}  & \textbf{21.44} \\ \bottomrule
\end{tabular}
\end{table}

\begin{figure}[!htbp]
\begin{center}
   \includegraphics[width=0.9\linewidth]{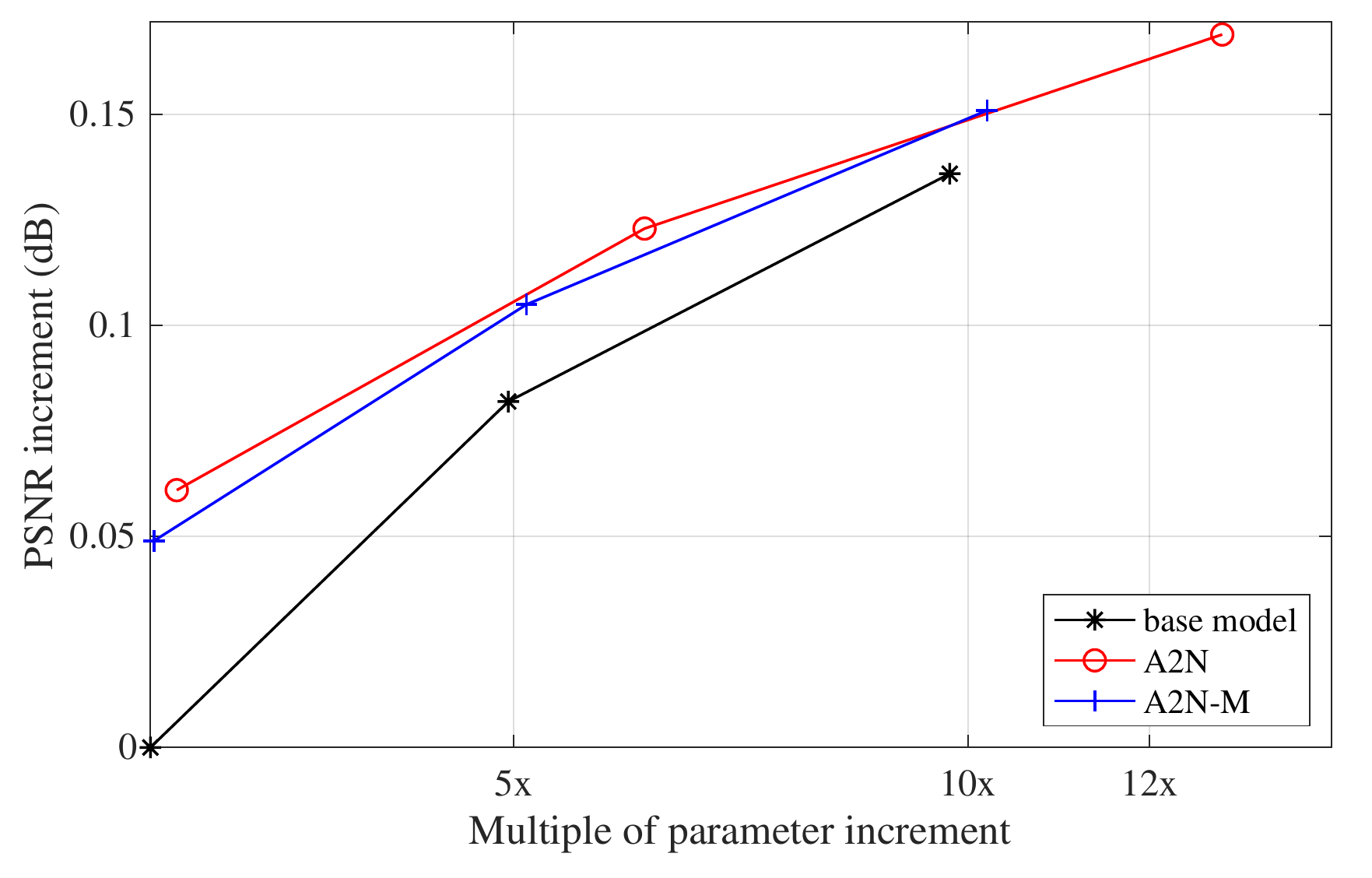}
\end{center}
\caption{Results of base model and our A$^2$ structure on different model size. Our method can get higher PSNR under the similar parameter number.}
\label{fig:size}
\end{figure}

\textbf{Comparison with Other Fusion Methods.}
Most SR models fuse features by addition \cite{kim2018ram} or concatenation \cite{hui2019lightweight,zhao2020efficient}. Some methods \cite{kim2016deeply,wang2020lightweight,wang2019lightweight} give adaptive weights to each feature, which means the independent weights will be learned automatically when training the model.
To demonstrate the effectiveness of our method,
we compare our dynamic attention module with other mainstream feature fusion methods: addition, concatenation, and adaptive weights.
\tablename~\ref{tab:ab} shows that within a similar parameter number, A$^2$ structure has a considerable improvement over other fusion methods.
If we reduce the channel number to 32, even with about 400K fewer parameters, 
A$^2$N can still obtain a better result than other fusion Methods.
It demonstrates that the dynamic attention module is a better feature fusion method than other methods.

\textbf{Results of Local Attribution Maps.}
\label{local_attribution_maps}
The proposed method can make good use of the information in the input LR image.
Recently, Gu \textit{et al} \cite{gu2021interpreting} propose a novel attribution approach called local attribution map (LAM), which performs attribution analysis of SR networks and aims at finding the input pixels that strongly influence the SR results.
The results are summarised to diffusion index (DI), an evaluation metric that measures  the ability of extraction and utilization of the information in the LR image.
A larger DI indicates more pixels are involved.
LAM highlights the pixels which have the greatest impact on the SR results. 
For the same local patch, if the LAM map involves more pixels or a larger range, it can be considered that the SR network has extracted and used the information from more pixels. 
We follow the suggested setting, and in \tablename~\ref{tab:lam} we show the PSNR and DI performances for some SR networks test on these images.
Among all the models, A$^2$N has the highest DI and PSNR.
We can notice that using the dynamic attention module makes DI much higher than other models.
\figurename~\ref{fig:lam} shows the LAM results, which visualize the importance of pixels.
The LAM results indicate that CARN and channel-spatial attention model only utilize very limited information, 
Our A$^2$N can utilize a wider range of information for better SR results for models without downsampling.
%

\begin{table}[h]
\renewcommand\arraystretch{1.1}
\begin{center}
\caption{Results of base model and our A$^2$ structure on different model size. The baseline model denotes models with only one attention trunk. Test on Set14 ($\times$ 4).}
\label{tab:size}
\begin{tabular}{ccll}
\toprule
\#channel           & \#block             & Model      & PSNR            \\ \midrule
\multirow{3}{*}{40} & \multirow{3}{*}{16} & baseline   & 28.646          \\
                    &                     & A$^2$N-M   & 28.695 (+0.049) \\ 
                    &                     & A$^2$N     & 28.707 (+0.061) \\ \midrule
\multirow{3}{*}{64} & \multirow{3}{*}{32} & baseline   & 28.728 (+0.082) \\
                    &                     & A$^2$N-M   & 28.751 (+0.105) \\ 
                    &                     & A$^2$N     & 28.769 (+0.123) \\ \midrule
\multirow{3}{*}{64} & \multirow{3}{*}{64} & baseline   & 28.782 (+0.136) \\
                    &                     & A$^2$N-M   & 28.797 (+0.151) \\ 
                    &                     & A$^2$N     & 28.815 (+0.169) \\ \bottomrule
\end{tabular}
\end{center}
\end{table}

\subsection{Ablation Study}

\textbf{Choice of channel width in Non-attention Branch.}
In A$^2$N, we use 3 $\times$ 3 convolution to extract features, in A$^2$N-M, 1 $\times$ 1 convolution is used.
We compare the results of 1 $\times$ 1 convolution, 3 $\times$ 3 convolution, and pass-through. 
From our experimental results shown in \tablename~\ref{tab:non-att}, 
even without any operations, A$^2$ structure is still better than the one-trunk attention structure.
Compare with the one-trunk structure and 1 $\times$ 1 convolution, 
1 $\times$ 1 convolution gain 0.049 dB improvement with only 33K parameters. 
It achieves a great trade-off between parameter number and performance.
The results of w/o operation and 1 $\times$ 1 convolution also prove that convolution in non-attention branch truly contributes to the network, 
it can extract the effective features that the attention branch cannot extract, 
so that it can complement the features of attention branch.
These comparisons firmly demonstrate the effectiveness of A$^2$ structure.


\begin{table*}[]
\renewcommand\arraystretch{1.1}
\begin{center}
\caption{Quantitative results (PSNR/SSIM) of state-of-the-art SR methods for all upscaling factors $\times$2, $\times$3, and $\times$4. {\color[HTML]{FE0000} Red}/{\color[HTML]{3531FF} Blue} text: best/second-best among all methods (excluding models with parameters greater than 2M).}
\label{tab:sota}
\begin{tabular}{cclccccccc}
\toprule
Scale                & Size Scope                         & \multicolumn{1}{c}{Model} & Params & MutiAdds & Set5                                & Set14                               & B100                                & Urban100                            & Manga109                            \\ \midrule
                     &                                    & FSRCANN \cite{dong2016accelerating}                    & 0.01M  & 6G       & 37.00/0.9558                        & 32.63/0.9088                        & 31.53/0.8920                        & 29.88/0.9020                        & 36.67/0.9710                        \\
                     &                                    & SRCNN \cite{dong2015image}                      & 0.06M  & 52.7G    & 36.66/0.9542                        & 32.45/0.9067                        & 31.36/0.8879                        & 29.50/0.8946                        & 35.60/0.9663                        \\
                     &                                    & DRRN \cite{tai2017image}                       & 0.3M   & 6797G    & 37.74/0.9591                        & 33.23/0.9136                        & 32.05/0.8973                        & 31.23/0.9188                        & 37.92/0.9760                        \\
                     &                                    & VDSR \cite{kim2016accurate}                       & 0.7M   & 612.6G   & 37.53/0.9587                        & 33.03/0.9124                        & 31.90/0.8960                        & 30.76/0.9140                        & 37.22/0.9729                        \\
                     &                                    & MemNet \cite{tai2017memnet}                    & 0.7M   & 2662.4G  & 37.78/0.9597                        & 33.28/0.9142                        & 32.08/0.8978                        & 31.31/0.9195                        & -                                   \\
                     &                                    & IMDN \cite{hui2019lightweight}                      & 0.7M   & 158.8G   & 38.00/0.9605                        & 33.63/0.9177                        & 32.19/0.8996                        & 32.17/0.9283                        & {\color[HTML]{FE0000} 38.88}/{\color[HTML]{FE0000} 0.9774}                        \\
                     &                                    & A2N-M (Ours)               & 0.8M   & 200.3G   & {\color[HTML]{FE0000} 38.06}/0.9601 & {\color[HTML]{3531FF} 33.73}/{\color[HTML]{3531FF} 0.9190} & {\color[HTML]{3531FF} 32.22}/0.8997 & {\color[HTML]{3531FF} 32.34}/{\color[HTML]{3531FF} 0.9300} & 38.80/0.9765 \\
                     &                                    & MADNet-$L_F$ \cite{lan2020madnet}              & 0.9M   & 187.1G   & 37.85/0.9600                        & 33.39/0.9161                        & 32.05/0.8981                        & 31.59/0.9234                        & -                                   \\
                     & \multirow{-9}{*}{\textless 1M}     & A2F-M \cite{wang2020lightweight}                     & 1M     & 224.2G   & 38.04/{\color[HTML]{3531FF} 0.9607} & 33.67/0.9184                        & 32.18/0.8996                        & 32.27/0.9294                        & {\color[HTML]{3531FF} 38.87}/{\color[HTML]{FE0000} 0.9774}                        \\ \cmidrule{2-10} 
                     &                                    & A2N (Ours)                 & 1M     & 247.5G   & {\color[HTML]{FE0000} 38.06}/{\color[HTML]{FE0000} 0.9608} & {\color[HTML]{FE0000} 33.75}/{\color[HTML]{FE0000} 0.9194} & {\color[HTML]{FE0000} 32.22}/{\color[HTML]{FE0000} 0.9002} & {\color[HTML]{FE0000} 32.43}/{\color[HTML]{FE0000} 0.9311} & {\color[HTML]{3531FF} 38.87}/0.9769 \\
                     &                                    & AWSRN-M \cite{wang2019lightweight}                    & 1M     & 244.1G   & 38.04/0.9605                        & 33.66/0.9181                        & 32.21/{\color[HTML]{3531FF} 0.9000} & 32.23/0.9294                        & 38.66/0.9772                        \\
                     &                                    & SRMDNF \cite{zhang2018learning}                     & 1.5M   & 347.7G   & 37.79/0.9600                        & 33.32/0.9150                        & 32.05/0.8980                        & 31.33/0.9200                        & -                                   \\
                     &                                    & CARN\cite{ahn2018fast}                       & 1.6M   & 222.8G   & 37.76/0.9590                        & 33.52/0.9166                        & 32.09/0.8978                        & 31.92/0.9256                        & -                                   \\
                     & \multirow{-5}{*}{\textless 2M}     & DRCN \cite{kim2016deeply}                       & 1.8M   & 17974G   & 37.63/0.9588                        & 33.04/0.9118                        & 31.85/0.8942                        & 30.75/0.9133                        & 37.63/0.9723                        \\ \cmidrule{2-10} 
                     &                                    & ERN \cite{9007041}                        & 9.5M   & -        & 38.18/0.9610                        & 33.88/0.9195                        & 32.30/0.9011                        & 32.66/0.9332                        & -                                   \\
                     &                                    & RCAN \cite{zhang2018image}                       & 16M    & -        & 38.27/0.9614                        & 34.12/0.9216                        & 32.41/0.9027                        & 33.34/0.9384                        & 39.44/0.9786                        \\
\multirow{-17}{*}{2} & \multirow{-3}{*}{\textgreater 9M} & EDSR \cite{lim2017enhanced}                       & 40.7M  & -        & 38.11/0.9602                        & 33.92/0.9195                        & 32.32/0.9013                        & 32.93/0.9351                        & 39.10/0.9773                        \\ \midrule
                     &                                    & FSRCANN \cite{dong2016accelerating}                    & 0.01M  & 6G       & 37.00/0.9558                        & 32.63/0.9088                        & 31.53/0.8920                        & 29.88/0.9020                        & 36.67/0.9710                        \\
                     &                                    & SRCNN \cite{dong2015image}                     & 0.06M  & 52.7G    & 32.75/0.9090                        & 29.28/0.8209                        & 28.41/0.7863                        & 26.24/0.7989                        & 30.59/0.9107                        \\
                     &                                    & DRRN \cite{tai2017image}                       & 0.3M   & 6797G    & 34.03/0.9244                        & 29.96/0.8349                        & 28.95/0.8004                        & 27.53/0.8378                        & 32.74/0.9390                        \\
                     &                                    & VDSR \cite{kim2016accurate}                      & 0.7M   & 612.6G   & 33.66/0.9213                        & 29.77/0.8314                        & 28.82/0.7976                        & 27.14/0.8279                        & 32.01/0.9310                        \\
                     &                                    & MemNet \cite{tai2017memnet}                    & 0.7M   & 2662.4G  & 34.09/0.9248                        & 30.00/0.8350                        & 28.96/0.8001                        & 27.56/0.8376                        & -                                   \\
                     &                                    & IMDN \cite{hui2019lightweight}                      & 0.7M   & 71.5G    & 34.36/0.9270                        & 30.32/0.8417                        & 29.09/0.8046                        & 28.17/0.8519                        & 33.61/0.9445                        \\
                     &                                    & A2N-M (Ours)               & 0.8M   & 96.6G    & {\color[HTML]{FE0000} 34.50}/{\color[HTML]{FE0000} 0.9279} & {\color[HTML]{3531FF} 30.41}/{\color[HTML]{FE0000} 0.8438} & {\color[HTML]{3531FF} 29.13}/0.8058 & {\color[HTML]{3531FF} 28.35}/{\color[HTML]{3531FF} 0.8563} & {\color[HTML]{FE0000} 33.79}/{\color[HTML]{FE0000} 0.9458} \\
                     & \multirow{-8}{*}{\textless 1M}     & MADNet-$L_F$ \cite{lan2020madnet}              & 0.9M   & 88.4G    & 34.14/0.9251                        & 30.20/0.8395                        & 28.98/0.8023                        & 27.78/0.8439                        & -                                   \\ \cmidrule{2-10} 
                     &                                    & A2F-M \cite{wang2020lightweight}                     & 1M     & 100G     & {\color[HTML]{FE0000} 34.50}/0.9278 & 30.39/0.8427                        & 29.11/0.8054                        & 28.28/0.8546                        & 33.66/0.9453                        \\
                     &                                    & A2N (Ours)                 & 1M     & 117.5G   & 34.47/{\color[HTML]{FE0000} 0.9279} & {\color[HTML]{FE0000} 30.44}/{\color[HTML]{3531FF} 0.8437} & {\color[HTML]{FE0000} 29.14}/{\color[HTML]{FE0000} 0.8059} & {\color[HTML]{FE0000} 28.41}/{\color[HTML]{FE0000} 0.8570} & {\color[HTML]{3531FF} 33.78}/{\color[HTML]{FE0000} 0.9458} \\
                     &                                    & AWSRN-M \cite{wang2019lightweight}  & 1.1M   & 116.6G   & 34.42/0.9275                        & 30.32/0.8419                        & {\color[HTML]{3531FF} 29.13}/{\color[HTML]{FE0000} 0.8059}                & 28.26/0.8545                        & 33.64/0.9450                        \\
                     &                                    & SRMDNF \cite{zhang2018learning}                     & 1.5M   & 156.3G   & 34.12/0.9250                        & 30.04/0.8370                        & 28.97/0.8030                        & 27.57/0.8400                        & -                                   \\
                     &                                    & CARN\cite{ahn2018fast}                       & 1.6M   & 118.8G   & 34.29/0.9255                        & 30.29/0.8407                        & 29.06/0.8034                        & 28.06/0.8493                        & -                                   \\
                     & \multirow{-6}{*}{\textless 2M}     & DRCN \cite{kim2016deeply}                       & 1.8M   & 17974G   & 33.82/0.9226                        & 29.76/0.8311                        & 28.80/0.7963                        & 27.15/0.8276                        & 32.31/0.9328                        \\ \cmidrule{2-10} 
                     &                                    & ERN \cite{9007041}                        & 9.5M   & -        & 34.62/0.9285                        & 30.51/0.8450                        & 29.21/0.8080                        & 28.61/0.8614                        & -                                   \\
                     &                                    & RCAN \cite{zhang2018image}                       & 16M    & -        & 34.74/0.9299                        & 30.65/0.8482                        & 29.32/0.8111                        & 29.09/0.8702                        & 34.44/0.9499                        \\
\multirow{-17}{*}{3} & \multirow{-3}{*}{\textgreater 9M}  & EDSR \cite{lim2017enhanced}                       & 43.1M  & -        & 34.65/0.9280                        & 30.52/0.8462                        & 29.25/0.8093                        & 28.80/0.8653                        & 34.17/0.9476                        \\ \midrule
                     &                                    & FSRCANN \cite{dong2016accelerating}                    & 0.01M  & 4.6G     & 30.71/0.8657                        & 27.59/0.7535                        & 26.98/0.7150                        & 24.62/0.7280                        & 27.90/0.8517                        \\
                     &                                    & SRCNN \cite{dong2015image}                     & 0.06M  & 52.7G    & 30.48/0.8628                        & 27.49/0.7503                        & 26.90/0.7101                        & 24.52/0.7221                        & 27.66/0.8505                        \\
                     &                                    & DRRN \cite{tai2017image}                       & 0.3M   & 6797G    & 31.68/0.8888                        & 28.21/0.7720                        & 27.38/0.7284                        & 25.44/0.7638                        & 29.46/0.8960                        \\
                     &                                    & VDSR \cite{kim2016accurate}                      & 0.7M   & 612.6G   & 31.35/0.8838                        & 28.01/0.7674                        & 27.29/0.7251                        & 25.18/0.7524                        & 28.83/0.8809                        \\
                     &                                    & MemNet \cite{tai2017memnet}                    & 0.7M   & 2662.4G  & 31.74/0.8893                        & 28.26/0.7723                        & 27.40/0.7281                        & 25.50/0.7630                        & -                                   \\
                     &                                    & IMDN \cite{hui2019lightweight}                      & 0.7M   & 40.9G    & 32.21/0.8948                        & 28.58/0.7811                        & 27.56/0.7353                        & 26.04/0.7838                        & 30.45/0.9075                        \\
                     & \multirow{-7}{*}{\textless 1M}     & A2N-M (Ours)               & 0.8M   & 60.6G    & 32.27/{\color[HTML]{3531FF} 0.8963} & {\color[HTML]{3531FF} 28.70}/{\color[HTML]{FE0000} 0.7842} & {\color[HTML]{FE0000} 27.61}/{\color[HTML]{FE0000} 0.7376} & {\color[HTML]{FE0000} 26.28}/{\color[HTML]{3531FF} 0.7919} & {\color[HTML]{3531FF} 30.59}/{\color[HTML]{3531FF} 0.9103} \\ \cmidrule{2-10} 
                     &                                    & MADNet-$L_F$ \cite{lan2020madnet}              & 1M     & 54.1G    & 32.01/0.8925                        & 28.45/0.7781                        & 27.47/0.7327                        & 25.77/0.7751                        & -                                   \\
                     &                                    & A2F-M \cite{wang2020lightweight}                     & 1M     & 56.7G    & {\color[HTML]{3531FF} 32.28}/0.8955                        & 28.62/0.7828                        & 27.58/0.7364                        & 26.17/0.7892                        & 30.57/0.9100                        \\
                     &                                    & A2N (Ours)                 & 1M     & 72.4G    & {\color[HTML]{FE0000} 32.30}/{\color[HTML]{FE0000}0.8966} & {\color[HTML]{FE0000} 28.71}/{\color[HTML]{FE0000} 0.7842} & {\color[HTML]{FE0000} 27.61}/{\color[HTML]{3531FF} 0.7374} & {\color[HTML]{3531FF} 26.27}/{\color[HTML]{FE0000} 0.7920} & {\color[HTML]{FE0000} 30.67}/{\color[HTML]{FE0000} 0.9110} \\
                     &                                    & AWSRN-M \cite{wang2019lightweight}                    & 1.3M   & 72G      & 32.21/0.8954                        & 28.65/0.7832                        & 27.60/0.7368                        & 26.15/0.7884                        & 30.56/0.9093                        \\
                     &                                    & SRMDNF \cite{zhang2018learning}                     & 1.6M   & 89.3G    & 31.96/0.8930                        & 28.35/0.7770                        & 27.49/0.7340                        & 25.68/0.7730                        & -                                   \\
                     &                                    & CARN\cite{ahn2018fast}                       & 1.6M   & 90.9G    & 32.13/0.8937                        & 28.60/0.7806                        & 27.58/0.7349                        & 26.07/0.7837                        & -                                   \\
                     & \multirow{-7}{*}{\textless 2M}     & DRCN \cite{kim2016deeply}                       & 1.8M   & 17974G   & 31.53/0.8854                        & 28.02/0.7670                        & 27.23/0.7233                        & 25.14/0.7510                        & 28.98/0.8816                        \\ \cmidrule{2-10} 
                     &                                    & ERN \cite{9007041}                        & 9.5M   & -        & 32.39/0.8975                        & 28.75/0.7853                        & 27.70/0.7398                        & 26.43/0.7966                        & -                                   \\
                     &                                    & RCAN \cite{zhang2018image}                       & 16M    & -        & 32.63/0.9002                        & 28.87/0.7889                        & 27.77/0.7436                        & 26.82/0.8087                        & 31.22/0.9173                        \\
\multirow{-17}{*}{4} & \multirow{-3}{*}{\textgreater 9M}  & EDSR \cite{lim2017enhanced}                       & 43.7M  & -        & 32.46/0.8968                        & 28.80/0.7876                        & 27.71/0.7420                        & 26.64/0.8033                        & 31.02/0.9148                        \\ \bottomrule
\end{tabular}
\end{center}
\end{table*}

\textbf{Results on Large Model.}
We conduct experiments on three different sizes of models to analyse the effectiveness of our A$^2$ structure. 
We chose the model which only keeps the attention branch to compare with A$^2$N.
We change the number of channels and A$^2$B and get three sizes of models: large, medium, and small.
From our experimental results shown in \tablename~\ref{tab:size} and \figurename~\ref{fig:size}, 
smaller models with A$^2$ structure will gain more bonus than larger models, since for large models the PSNR tends to be saturated. 
A$^2$ structure could also conspicuously improve their performance for large models.
Note that the cost of A$^2$ structure is almost negligible, these improvements are  outstanding. 
From \figurename~\ref{fig:size}, our method can always get higher PSNR under a similar parameter number.

\section{Conclusions}
             
In this work, we show that not all attention modules are equally beneficial, the use of some attention modules can even cause a performance drop. The findings can help later people better use and understand attention in image super-resolution tasks. 
We than propose attention in attention network (A$^2$N) and building block A$^2$B for image SR. 
A$^2$ structure allows for more aggressive pixel-wise attention adjustments and could dynamically adjust the contribution of attention layers, allowing them to be penalized less frequently. 
%
%
Experiments have demonstrated that our method could achieve superior performances compared with state-of-the-art SR models of similar sizes.

\newpage

\section{Biography Section}
 




\begin{IEEEbiographynophoto}{Haoyu Chen}
 received his B Eng. degree in computer science and engineering from the Chinese University of Hong Kong, Shenzhen, in 2021. His research interests include computer vision, image processing and image generation.\end{IEEEbiographynophoto}

\begin{IEEEbiography}[{\includegraphics[width=1in,height=1.25in,clip,keepaspectratio]{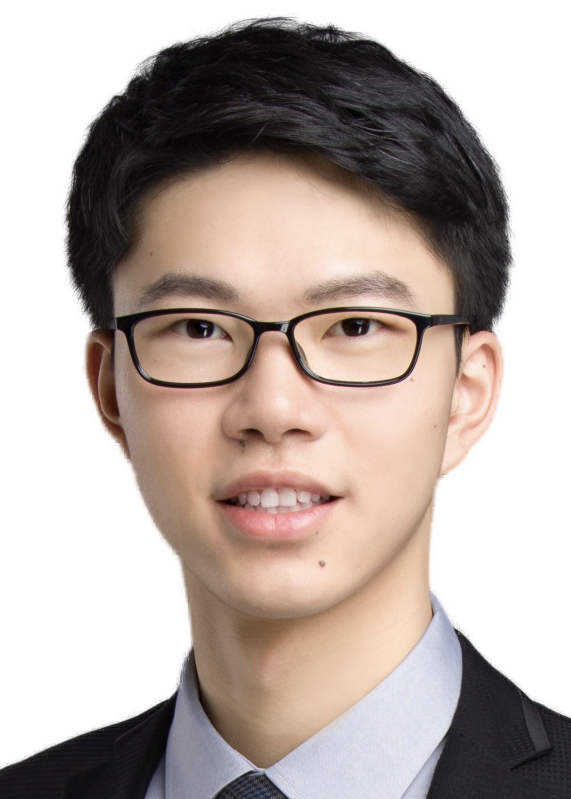}}]{Jinjin Gu}
received his B.Eng. degree in computer science and engineering from the Chinese University of Hong Kong, Shenzhen, in 2020.
He is currently pursuing a Ph.D. degree in Engineering and IT with the University of Sydney.
His research interests include computer vision, image processing, interpretability of deep learning algorithms and the applications of machine learning in industrial.
\end{IEEEbiography}

\begin{IEEEbiography}[{\includegraphics[width=1in,height=1.25in,clip,keepaspectratio]{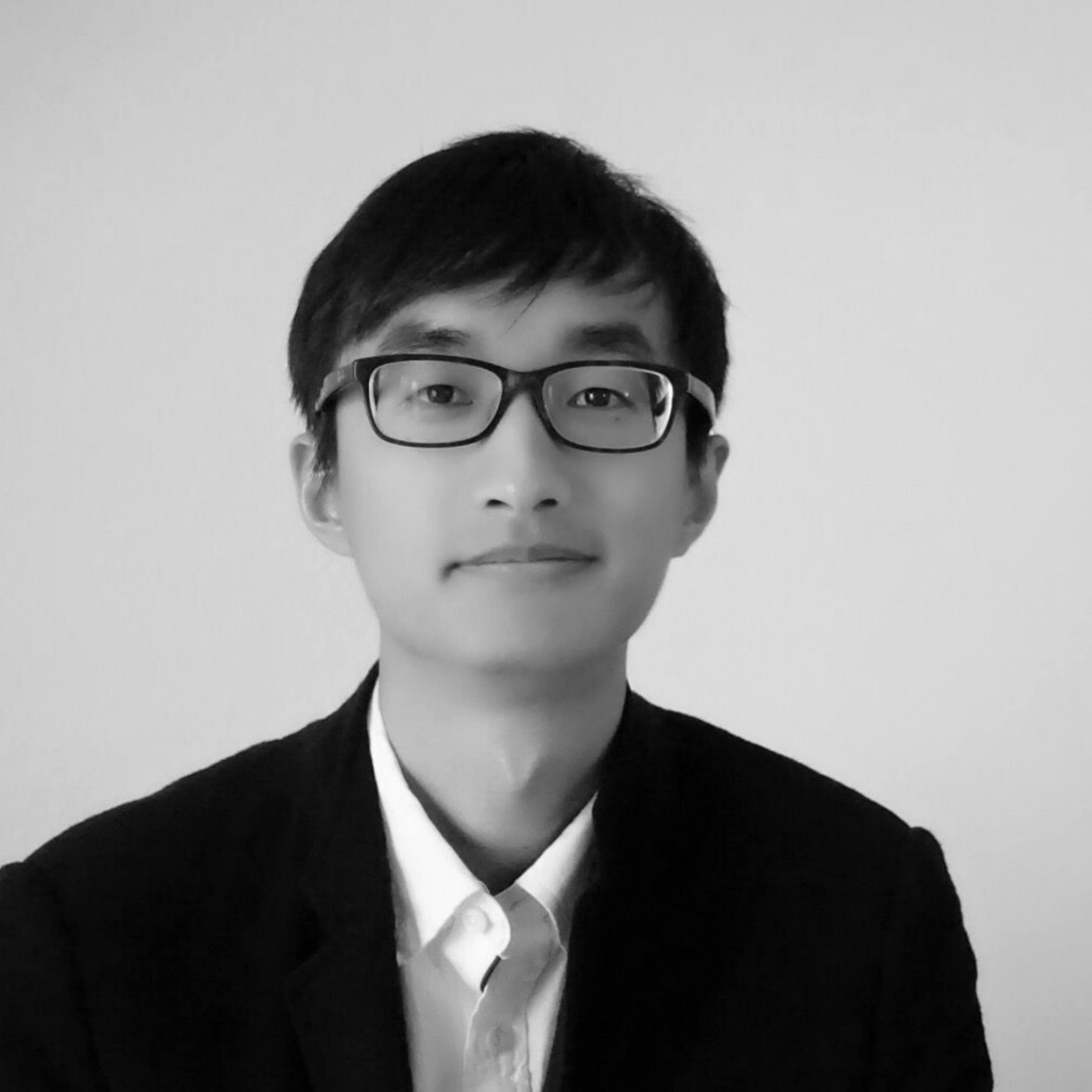}}]{Zhi Zhang}
 received his B.S. degree in electronic and information technology from Beijing Jiaotong University, Beijing, China, in 2012, and the M.S. and Ph.D. degrees in computer engineering from the University of Missouri-Columbia, Columbia, MO, USA, in 2014 and 2018, respectively. He joined Amazon AI in 2018 in pursuing highly efficient deep learning frameworks and toolkits. His current research interests include object detection, segmentation, and deep network acceleration.
\end{IEEEbiography}

\vfill

\end{document}